# Solve fraud detection problem by using graph based learning methods

Loc Tran[1], Tuan Tran[2], Linh Tran[3] and An Mai[4]

[1]John von Neumann Institute, VNU-HCM, Ho Chi Minh City, Vietnam (E-mail: tran0398@umn.edu)
[2]John von Neumann Institute, VNU-HCM, Ho Chi Minh City, Vietnam (E-mail: tuan.tran2016@qcf.jvn.edu.vn)
[3]Thu Dau Mot University, Binh Duong Province, Vietnam (E-mail: linhtran.cntt@tdmu.edu.vn)
[4]John von Neumann Institute, VNU-HCM, Ho Chi Minh City, Vietnam (E-mail: an.mai@jvn.edu.vn)

**Abstract**

The credit cards' fraud transactions detection is the important problem in machine learning field. To detect the credit cards' fraud transactions help reduce the significant loss of the credit cards' holders and the banks. To detect the credit cards' fraud transactions, data scientists normally employ the un-supervised learning techniques and supervised learning technique. In this paper, we employ the graph p-Laplacian based semi-supervised learning methods combined with the under-sampling technique such as Cluster Centroids to solve the credit cards' fraud transactions detection problem. Experimental results show that that the graph p-Laplacian semi-supervised learning methods outperform the current state of art graph Laplacian based semi-supervised learning method ($p=2$).

**2010 AMS Classification: 05C85**

**Article type: Research article**

**1 Introduction**

While purchasing online, the transactions can be done by using credit cards that are issued by the bank. In this case, if the cards or cards' details are stolen, the fraud transactions can be easily carried out. This will lead to the significant loss of the card holder or the bank. In order to detect credit cards' fraud transactions, data scientists employ a lot of machine learning techniques. To the best of our knowledge, there are two classes of machine learning techniques used to detect credit cards' fraud transactions which are un-supervised learning techniques and supervised learning techniques. The un-supervised learning techniques used to detect credit cards' fraud transactions are k-means clustering technique [1], k-nearest neighbors technique [1], Local Outlier Factor technique [1], to name a few. The supervised learning techniques used to detect credit cards' fraud transactions are Hidden Markov Model technique [2], neural network technique [3], Support Vector Machine technique [4], to name a few.

To the best of our knowledge, the graph based semi-supervised learning techniques [5] have not been applied to the credit cards' fraud transactions detection problem. In this paper, we will apply the un-normalized graph p-Laplacian based semi-supervised learning technique [6, 7] combined with the under-sampling technique to the credit cards' fraud transactions detection problem.

We will organize the paper as follows: Section 2 will introduce the preliminary notations and definitions used in this paper. Section 3 will introduce the definitions of the gradient and divergence operators of graphs. Section 4 will introduce the definition of Laplace operator of graphs and its properties. Section 5 will introduce the definition of the curvature operator of graphs and its properties. Section 6 will introduce the definition of the p-Laplace operator of graphs and its properties. Section 7 will show how to derive the algorithm of the un-normalized graph p-Laplacian based semi-supervised learning method from regularization framework. In section 8, we will compare the accuracy performance measures of the un-normalized graph Laplacian based semi-supervised learning algorithm (i.e. the current state of art graph based semi-supervised learning method) combined with the under-sampling technique such as Cluster Centroids technique [8] and the un-normalized graph p-Laplacian based semi-supervised learning algorithms combined with Cluster Centroids technique [8]. Section 9 will conclude this paper and the future direction of researches will be discussed.

**2 Preliminary notations and definitions**

Given a graph $G=(V,E,W)$ where $V$ is a set of vertices with $|V| = n$, $E \subseteq V * V$ is a set of edges and $W$ is a $n * n$ similarity matrix with elements $w_{ij} \geq 0$ ($1 \leq i,j \leq n$).

Also, please note that $w_{ij} = w_{ji}$.

The degree function $d: V \to R^+$ is

$$d_i = \sum_{j \sim i} w_{ij}, \quad (1)$$

where $j \sim i$ is the set of vertices adjacent with $i$.

Define $D = diag(d_1, d_2, \ldots, d_n)$.

The inner product on the function space $R^V$ is

$$<f, g>_V = \sum_{i \in V} f_i g_i \quad (2)$$

Also define an inner product on the space of functions $R^E$ on the edges

$$<F, G>_E = \sum_{(i,j) \in E} F_{ij} G_{ij} \quad (3)$$

Here let $H(V) = (R^V, <.,.>_V)$ and $H(E) = (R^E, <.,.>_E)$ be the Hilbert space real-valued functions defined on the vertices of the graph $G$ and the Hilbert space of real-valued functions defined in the edges of $G$ respectively.

## 3 Gradient and Divergence Operators

We define the gradient operator $d: H(V) \to H(E)$ to be

$$(df)_{ij} = \sqrt{w_{ij}}(f_j - f_i), \quad (4)$$

where $f: V \to R$ be a function of $H(V)$.

We define the divergence operator $div: H(E) \to H(V)$ to be

$$<df, F>_{H(E)} = <f, -divF>_{H(V)}, \quad (5)$$

where $f \in H(V), F \in H(E)$.

Thus, we have

$$(divF)_j = \sum_{i \sim j} \sqrt{w_{ij}} (F_{ji} - F_{ij}) \quad (6)$$

## 4 Laplace operator

We define the Laplace operator $\Delta: H(V) \to H(V)$ to be

$$\Delta f = -\frac{1}{2} div(df) \quad (7)$$

Thus, we have

$$(\Delta f)_j = d_j f_j - \sum_{i \sim j} w_{ij} f_i \quad (8)$$

The graph Laplacian is a linear operator. Furthermore, the graph Laplacian is self-adjoint and positive semi-definite.

Let $S_2(f) = <\Delta f, f>$, we have the following **theorem 1**

$$D_f S_2 = 2\Delta f \quad (9)$$

The proof of the above theorem can be found from [6, 7].

## 5 Curvature operator

We define the curvature operator $\kappa: H(V) \to H(V)$ to be

$$\kappa f = -\frac{1}{2} div(\frac{df}{\|df\|}) \quad (10)$$

Thus, we have

$$(\kappa f)_j = \frac{1}{2}\Sigma_{i \sim j} w_{ij} (\frac{1}{\|d_i f\|} + \frac{1}{\|d_j f\|})(f_j - f_i) \quad (11)$$

From the above formula, we have

$$d_i f = ((df)_{ij} : j \sim i)^T \quad (12)$$

The local variation of $f$ at $i$ is defined to be

$$\|d_i f\| = \sqrt{\Sigma_{j \sim i}(df)_{ij}^2} = \sqrt{\Sigma_{j \sim i} w_{ij}(f_j - f_i)^2} \quad (13)$$

To avoid the zero denominators in (11), the local variation of $f$ at $i$ is defined to be

$$\|d_i f\| = \sqrt{\Sigma_{j \sim i}(df)_{ij}^2 + \epsilon}, \quad (14)$$

where $\epsilon = 10^{-10}$.

The graph curvature is a non-linear operator.

Let $S_1(f) = \Sigma_i \|d_i f\|$, we have the following **theorem 2**

$$D_f S_1 = \kappa f \quad (15)$$

The proof of the above theorem can be found from [6, 7].

## 6 p-Laplace operator

We define the p-Laplace operator $\Delta_p: H(V) \to H(V)$ to be

$$\Delta_p f = -\frac{1}{2} div(\|df\|^{p-2} df) \quad (16)$$

Thus, we have

$$(\Delta_p f)_j = \frac{1}{2} \Sigma_{i \sim j} w_{ij} (\|d_i f\|^{p-2} + \|d_j f\|^{p-2})(f_j - f_i) \quad (17)$$

Let $S_p(f) = \frac{1}{p} \Sigma_i \|d_i f\|^p$, we have the following **theorem 3**

$$D_f S_p = p \Delta_p f \quad (18)$$

**7 Discrete regularization on graphs and credit cards' fraud transactions detection problems**

Given a transaction network $G=(V,E)$. $V$ is the set of all transactions in the network and $E$ is the set of all possible interactions between these transactions. Let $y$ denote the initial function in $H(V)$. $y_i$ can be defined as follows

$$y_i = \begin{cases} 1 \text{ if transaction } i \text{ is the fraud transaction} \\ -1 \text{ if transaction } i \text{ is the normal transaction} \\ 0 \text{ otherwise} \end{cases}$$

Our goal is to look for an estimated function $f$ in $H(V)$ such that $f$ is not only smooth on $G$ but also close enough to an initial function $y$. Then each transaction $i$ is classified as $sign(f_i)$. This concept can be formulated as the following optimization problem

$$argmin_{f \in H(V)} \{S_p(f) + \frac{\mu}{2} \|f - y\|^2\} \quad (19)$$

The first term in (19) is the smoothness term. The second term is the fitting term. A positive parameter $\mu$ captures the trade-off between these two competing terms.

### 7.1) p-smoothness

For any number $p$, the optimization problem (19) is

$$argmin_{f \in H(V)} \{\frac{1}{p} \Sigma_i \|d_i f\|^p + \frac{\mu}{2} \|f - y\|^2\}, \quad (20)$$

By theorem 3, we have

**Theorem 4:** The solution of (20) satisfies

$$\Delta_p f + \mu(f - y) = 0, \quad (21)$$

The *p-Laplace* operator is a non-linear operator; hence we do not have the closed form solution of equation (21). Thus, we have to construct iterative algorithm to obtain the solution. From (21), we have

$$\frac{1}{2}\Sigma_{i\sim j} w_{ij} \left( \|d_i f\|^{p-2} + \|d_j f\|^{p-2} \right)(f_j - f_i) + \mu(f_j - y_j) = 0 \quad (22)$$

Define the function $m: E \to R$ by

$$m_{ij} = \frac{1}{2} w_{ij}(\|d_i f\|^{p-2} + \|d_j f\|^{p-2}) \quad (23)$$

Then equation (22) which is

$$\sum_{i\sim j} m_{ij}(f_j - f_i) + \mu(f_j - y_j) = 0$$

can be transformed into

$$\left(\Sigma_{i\sim j} m_{ij} + \mu\right) f_j = \Sigma_{i\sim j} m_{ij} f_i + \mu y_j \quad (24)$$

Define the function $p: E \to R$ by

$$p_{ij} = \begin{cases} \frac{m_{ij}}{\Sigma_{i\sim j} m_{ij} + \mu} & if\ i \neq j \\ \frac{\mu}{\Sigma_{i\sim j} m_{ij} + \mu} & if\ i = j \end{cases} \quad (25)$$

Then

$$f_j = \Sigma_{i\sim j} p_{ij} f_i + p_{jj} y_j \quad (26)$$

Thus we can consider the iteration

$$f_j^{(t+1)} = \Sigma_{i\sim j} p_{ij}^{(t)} f_i^{(t)} + p_{jj}^{(t)} y_j \text{ for all } j \in V$$

to obtain the solution of (20).

## 8 Experiments and results

### Datasets

In this paper, we use the transaction dataset available from [9]. This dataset contains 284,807 transactions. Each transaction has 30 features. In the other words, we are given transaction data matrix ($R^{284807*30}$) and the annotation (i.e. the label) matrix ($R^{284807*1}$). The ratio between the number of fraud transactions and the number of normal transactions is 0.00173. Hence we easily recognize that this is the imbalanced classification problem. In order to solve this imbalanced classification problem, we initially apply the under-sampling technique which is the Cluster Centroid technique [8] to this imbalanced dataset. Then we have that the ratio between the number of fraud transactions and the number of normal transactions is 0.4. In the other words, we are given the **new transaction data** matrix ($R^{1722*30}$) and the annotation (i.e. the label) matrix ($R^{1722*1}$).

Then we construct the similarity graph from the transaction data. The similarity graph used in this paper is the k-nearest neighbor graph: Transaction $i$ is connected with transaction $j$ if transaction $i$ is among the k-nearest neighbor of transaction $j$ or transaction $j$ is among the k-nearest neighbor of transaction $i$.

In this paper, the similarity function is the Gaussian similarity function

$$s(T(i,:),T(j,:)) = \exp(-\frac{d(T(i,:),T(j,:))}{t})$$

In this paper, $t$ is set to 0.1 and the 5-nearest neighbor graph is used to construct the similarity graph from the **new transaction data**.

## Experimental Results

In this section, we experiment with the above proposed un-normalized graph p-Laplacian methods with *p=1, 1.1, 1.2, 1.3, 1.4, 1.5, 1.6, 1.7, 1.8, 1.9* and the current state of the art method (i.e. the un-normalized graph Laplacian based semi-supervised learning method *p=2*) in terms of classification accuracy performance measure. The accuracy performance measure Q is given as follows

$$Q = \frac{True\ Positive + True\ Negative}{True\ Positive + True\ Negative + False\ Positive + False\ Negative}$$

All experiments were implemented in Matlab 6.5 on virtual machine. The **new transaction data** is divided into two subsets: the training set and the testing set. The training set contains 1,208 transactions. The testing set contains 514 transactions. The parameter $\mu$ is set to 1.

The accuracy performance measures of the above proposed methods and the current state of the art method is given in the following table 1

| Accuracy Performance Measures (%) | |
| --- | --- |
| p=1 | 88.52 |
| p=1.1 | 88.52 |
| p=1.2 | 88.52 |
| p=1.3 | 88.52 |
| p=1.4 | 88.52 |
| p=1.5 | 88.52 |
| p=1.6 | 88.52 |
| p=1.7 | 88.52 |
| p=1.8 | 88.52 |
| p=1.9 | 88.52 |
| p=2 | 88.33 |

Table 1: The comparison of accuracies of proposed methods with different p-values

From the above table, we easily recognized that the un-normalized graph p-Laplacian semi-supervised learning methods outperform the current state of art method. The results from the above table shows that the un-normalized graph p-Laplacian semi-supervised learning methods are at least as good as the current state of the art method (*p=2*) but often lead to better classification accuracy performance measures.

## 9 Conclusions

We have developed the detailed regularization frameworks for the un-normalized graph p-Laplacian semi-supervised learning methods applying to the credit cards' fraud transactions detection problem. Experiments show that the un-normalized graph p-Laplacian semi-supervised learning methods are at least as good as the current state of the art method (i.e. *p=2*) but often lead to significant better classification accuracy performance measures.

In the future, we will develop the detailed regularization frameworks for the un-normalized hypergraph p-Laplacian semi-supervised learning methods and will apply these methods to this credit cards' fraud transactions detection problem.

**Acknowledgement**

This research is funded by Vietnam National University Ho Chi Minh City (VNU-HCM) under grant number C2018-42-02.**References**

[1] Goldstein, Markus, and Seiichi Uchida. "A comparative evaluation of unsupervised anomaly detection algorithms for multivariate data." *PloS one* 11.4 (2016): e0152173.

[2] Singh, Anshul, and Devesh Narayan. "A survey on hidden markov model for credit card fraud detection." *International Journal of Engineering and Advanced Technology (IJEAT)* 1.3 (2012).

[3] Nune, Ganesh Kumar, and P. Vasanth Sena. "Novel Artificial Neural Networks and Logistic Approach for Detecting Credit Card Deceit." *International Journal of Computer Science and Network Security (IJCSNS)* 13.9 (2013): 58.

[4] Şahin, Yusuf G., and Ekrem Duman. "Detecting credit card fraud by decision trees and support vector machines." (2011).

[5] Shin, Hyunjung, Andreas Martin Lisewski, and Olivier Lichtarge. "Graph sharpening plus graph integration: a synergy that improves protein functional classification." *Bioinformatics* 23.23 (2007): 3217-3224.